\documentclass{article}

 \usepackage[preprint]{neurips_2026}

\usepackage[utf8]{inputenc} 
\usepackage[T1]{fontenc}    
\usepackage{hyperref}       
\usepackage{url}            
\usepackage{booktabs}       
\usepackage{amsfonts}       
\usepackage{nicefrac}       
\usepackage{microtype}      
\usepackage{xcolor}         
\usepackage{amsmath}        
\usepackage{cleveref}       
\usepackage{colortbl}       
\usepackage{array}          
\usepackage{graphicx}       
\usepackage{makecell}       
\usepackage{amssymb}        
\usepackage{multirow}
\usepackage[table]{xcolor}
\usepackage{pifont}         
\usepackage{adjustbox}
\usepackage{minitoc}
\usepackage{wrapfig}
\usepackage{caption}
\usepackage{subcaption}

\newcolumntype{O}{>{\columncolor{red!10}\centering\arraybackslash\bfseries}c}

\usepackage{MnSymbol,bbding,pifont}
\usepackage{tikz}
\usepackage{fontawesome5}
\definecolor{commentcolor}{RGB}{153, 0, 0}
\definecolor{electricindigo}{rgb}{0.44, 0.0, 1.0}
\definecolor{red}{rgb}{1, 0.0, 0.0}
\definecolor{ceruleanblue}{rgb}{0.16, 0.32, 0.75}
\definecolor{fireenginered}{rgb}{0.81, 0.09, 0.13}
\definecolor{olive}{rgb}{0.6, 0.6, 0.2}
\definecolor{sand}{rgb}{0.8666666666666667, 0.8, 0.4666666666666667}
\definecolor{wine}{rgb}{0.5333333333333333, 0.13333333333333333, 0.3333333333333333}
\definecolor{deblue}{RGB}{11,132,147}
\definecolor{ocra}{RGB}{204, 119, 34}

\definecolor{electricindigo}{rgb}{0.44, 0.0, 1.0}
\definecolor{lightblue}{RGB}{240,245,255}
\definecolor{darkblue}{RGB}{40,40,85}
\definecolor{babyblue}{rgb}{0.54, 0.81, 0.94}
\definecolor{pearDark}{HTML}{2980B9}
\definecolor{pearDarker}{HTML}{1D2DEC}

\newcommand{\cmark}{\textcolor{teal}{\ding{52}}}
\newcommand{\xmark}{\textcolor{wine}{\ding{56}}}

\title{Causal Spatiotemporal State Evolution  \\ for Monocular 4D Reconstruction}

\author{%
  Qiaowei Miao\thanks{Equal contribution.} \\
  Zhejiang University \\
  \And
  Yijing Guo\footnotemark[1] \\
  ShanghaiTech University \\
  \And
  Lipei Zhang\footnotemark[1] \\
  University of Cambridge \\
  \And
  Qifeng Cheng \\
  NIO \\
  \And
  Yuru Xiao \\
  Harbin Institute of Technology \\
  \And
  Ziqian Ni \\
  NIO \\
  \And
  Sheng Yang \\
  NIO \\
}

\begin{document}

\maketitle

\begin{abstract}

While feed-forward models excel in static 3D geometry estimation, extending them to dynamic 4D scenes requires modeling visual geometry as a causally evolving state. Current methods typically rely on unordered temporal aggregation, which lacks the directional structure of physical time. Furthermore, existing supervision often treats geometry and motion as disjoint objectives, failing to exploit their inherent physical coupling. Consequently, they lack a unified constraint to effectively supervise temporal state transitions, leading to inconsistent 4D scene representations.
To address this, we propose a feed-forward framework for causal 4D visual geometry learning from monocular videos. At its core, {Causal Spatiotemporal Mamba (CSM)} augments temporal aggregation with directional state propagation. Through complementary \textit{Temporal-Spatial} and \textit{Spatial-Temporal Scans}, CSM continuously propagates geometry-aware states across frames to model spatiotemporal evolution. To align this causal state evolution with observable dynamics, we introduce {Unified Motion Supervision (UMS)}. Rather than treating motion as an independent objective, UMS leverages geometry-aware predictions from the camera and depth heads to construct a unified forward motion field that is consistent with the underlying 3D structure.
This geometry-grounded supervision explicitly constrains inter-frame state transitions, enabling stable states propagation across frames. Extensive evaluations on dynamic benchmarks demonstrate state-of-the-art performance across video depth, camera pose, and optical flow estimation. These results validate that the synergy between the directional state propagation of Causal Spatiotemporal Mamba and Unified Motion Supervision effectively captures continuous scene evolution, significantly reducing temporal drift and reinforcing spatial-temporal consistency in 4D representation learning.

\end{abstract}

\section{Introduction}
\label{sec:intro}


While feed-forward models have driven remarkable progress in static 3D geometry estimation~\cite{wang2024dust3r, zhang2025monst3r, wang2025vggt, wang2025pi, wang2025moge, wang2025moge_v2}, extending this paradigm to dynamic 4D scenes~\cite{zhang2025monst3r, zhuo2026streamvggt, fang2026more, wang2025_4dvggt, cong2026flow3r} calls for modeling visual geometry as a temporally evolving state, rather than simply aggregating observations across frames. Unlike static reconstruction, which primarily relies on cross-view geometric consistency, 4D geometry learning must further account for temporally ordered state propagation and motion-consistent inter-frame constraints. However, existing feed-forward spatiotemporal models often rely on temporal aggregation mechanisms that weakly encode the directional structure of time, while geometry and motion are typically supervised through separate objectives. This separation limits the supervision of geometry-consistent state evolution and can lead to temporally unstable dynamic geometry estimates. These limitations motivate a causal spatiotemporal formulation in which geometry-aware states are propagated along the video order and guided by geometry-grounded motion constraints.


However, existing feed-forward 4D methods lack an explicit mechanism for modeling such causal temporal state evolution. 
The dominant paradigm extends static 3D backbones by applying attention-based temporal aggregation to video tokens~\cite{vaswani2017attention, wang2025vggt}. 
Although global or masked attention~\cite{zhuo2026streamvggt, fang2026more} can restrict the visible context, it primarily captures temporal interactions through pairwise token affinities rather than directional state propagation.
Consequently, these methods control which tokens are accessed, but do not explicitly define how geometry-aware states evolve sequentially, limiting their ability to model dynamic 4D scenes as causally ordered state transitions.


Furthermore, existing frameworks lack a unified supervisory formulation that couples geometry and motion during temporal state evolution.
Recent feed-forward 4D methods~\cite{zhang2025monst3r, zhuo2026streamvggt, fang2026more, wang2025_4dvggt, cong2026flow3r} typically introduce motion cues, such as optical flow, scene flow, or segmentation masks, as auxiliary output-level objectives.
Yet in dynamic 4D scenes, inter-frame motion is not an independent signal, but a geometry-dependent manifestation of the evolving scene state.
Treating motion supervision separately from geometry evolution therefore weakens its role as a structural constraint on temporal state transitions.
This motivates unified motion supervision that grounds inter-frame motion in evolving 3D geometry, encouraging more coherent 4D geometry estimation.

To overcome these limitations, we formulate monocular 4D visual geometry learning as directionally ordered spatiotemporal state evolution.
We instantiate this formulation with \textbf{Causal Spatiotemporal Mamba (CSM)}, a feed-forward framework that moves beyond correlation-based temporal aggregation by introducing directional state propagation over geometry-aware representations.
To capture complementary spatial and temporal dependencies, CSM introduces \textit{Temporal-Spatial Scan} and \textit{Spatial-Temporal Scan}, which propagate geometry-aware states along ordered video streams while respecting temporal causality.
To align this propagation with observable dynamics, we further introduce \textbf{Unified Motion Supervision (UMS)}.
UMS constructs a unified forward motion field from geometry-aware predictions, coupling inter-frame motion with evolving 3D geometry and providing a structural constraint for learning temporally coherent state transitions.
Our contributions are summarized as follows:
\begin{itemize}
    \item We formulate monocular 4D visual geometry learning as \textit{directionally ordered spatiotemporal state evolution}, in which geometry-aware states are modeled along the temporal order of video frames to capture dynamic geometry evolution.

    \item We propose \textbf{Causal Spatiotemporal Mamba (CSM)}, a feed-forward state-space framework that introduces complementary \textit{Temporal-Spatial Scan} and \textit{Spatial-Temporal Scan} to propagate geometry-aware states across video frames while respecting temporal causality.

    \item We introduce \textbf{Unified Motion Supervision (UMS)}, a geometry-grounded supervision mechanism that constructs a unified forward motion field from predicted geometry states, coupling inter-frame motion with evolving 3D geometry to guide temporally coherent state evolution.
\end{itemize}

Extensive experiments on multiple dynamic 4D reconstruction benchmarks demonstrate the effectiveness of our model, showing strong performance in dynamic geometry estimation.




\section{Related Work}
\label{sec:related}


\paragraph{Feed-forward 4D Visual Geometry.}
Feed-forward reconstruction methods have shifted geometry recovery from per-scene optimization to direct prediction, showing strong generalization in static 3D reconstruction~\cite{wang2024dust3r,wang2025vggt,wang2025pi,wu2025point3r}. 
Recent works extend this paradigm to dynamic 4D scenes through motion-aware geometry estimation~\cite{zhang2025monst3r}, streaming temporal processing~\cite{zhuo2026streamvggt}, motion-aware masking~\cite{fang2026more}, spatiotemporal geometry prediction~\cite{wang2025_4dvggt}, and flow- or scene-flow-based supervision~\cite{cong2026flow3r}. 
These methods introduce temporal context and motion cues into feed-forward reconstruction, but temporal reasoning is commonly realized through attention-based aggregation, streaming windows, or pairwise matching. 
Such designs model interactions among frames, yet do not explicitly define a directional update rule for propagating geometry-aware states along the video order. 
In addition, motion cues are often used as auxiliary flow, scene-flow, mask, or prior-based objectives, leaving their relation to continuous geometry evolution only implicitly modeled. 
Consequently, the coupling between inter-frame motion and geometry-consistent state transitions remains limited.
In contrast, our framework formulates monocular 4D visual geometry as causal spatiotemporal state evolution, instantiated by Causal Spatiotemporal Mamba (CSM) and guided by Unified Motion Supervision (UMS). 
We summarize these differences in Table~\ref{tab:feedforward-4d-comparison}.

\begin{table}[t]
\centering
\caption{\textbf{Methodological comparison of feed-forward 3D/4D visual geometry representations.} 
We compare spatiotemporal modeling paradigms and supervision mechanisms across feed-forward 4D methods. 
Our model combines temporally ordered state propagation with \textit{Unified Motion Supervision (UMS)}, which provides geometry-grounded supervision by coupling inter-frame motion with evolving 3D geometry, rather than treating motion cues as isolated auxiliary objectives.
}
\label{tab:feedforward-4d-comparison}
\adjustbox{max width=\textwidth}{%
\begin{tabular}{lccccc}
\toprule
\multirow{2}{*}{\textbf{Method}} &
\multicolumn{2}{c}{\textbf{Spatiotemporal Modeling}} &
\multicolumn{3}{c}{\textbf{Supervision}} \\
\cmidrule(lr){2-3}\cmidrule(lr){4-6}
&
\textbf{Global Correlation} &
\textbf{Causality Order} &
\textbf{Geometry} &
\textbf{Motion} &
\textbf{Unified Geo-Motion} \\
\midrule
VGGT~\cite{wang2025vggt}           & \cmark & \xmark & \cmark & \xmark & \xmark \\
CUT3R~\cite{wang2025cut3r}           & \cmark & \xmark & \cmark & \xmark & \xmark \\
StreamVGGT~\cite{zhuo2026streamvggt} & \cmark & \xmark & \cmark & \xmark & \xmark \\
MoRE~\cite{fang2026more}             & \cmark & \xmark & \cmark & \cmark & \xmark \\
PAGE-4D~\cite{zhou2026page4d}        & \cmark & \xmark & \cmark & \cmark & \xmark \\
Flow3r~\cite{cong2026flow3r}         & \cmark & \xmark & \cmark & \cmark & \xmark \\
\midrule
\rowcolor{gray!12}
\textbf{Ours}                  & \cmark & \cmark & \cmark & \cmark & \cmark \\
\bottomrule
\end{tabular}}
\vspace{-0.6em}
\end{table}

\paragraph{State Space Models for Visual Representation.}
Structured State Space Models (SSMs) provide an efficient alternative to attention for sequence modeling~\cite{gu2022efficiently}, and Mamba further enables input-dependent selective state updates~\cite{mamba,mamba2}. 
Recent works have applied Mamba-style modules to image and video representation learning~\cite{liu2024vmamba,li2024videomamba}, as well as static 3D reconstruction and novel view synthesis~\cite{shen2025gamba}. 
However, feed-forward 4D visual geometry requires more than directly applying generic visual scans to video tokens: the model needs to preserve spatial geometric structure while propagating states along the temporal order of dynamic scenes. 
Existing image- or video-oriented scan patterns are typically designed for general visual representation learning, rather than geometry-aware state evolution in dynamic 4D reconstruction. 
CSM addresses this gap with complementary \textit{Temporal-Spatial Scan} and \textit{Spatial-Temporal Scan}, which adapt state-space propagation to temporally ordered 4D geometry learning.


\paragraph{Motion Supervision for Dynamic Geometry.}
Cross-frame motion correspondences provide important supervision for learning dynamic geometry. 
Point tracking methods such as TAPIR~\cite{doersch2023tapir} and CoTracker3~\cite{karaev2024cotracker3} offer reliable but sparse matches, while optical flow methods such as RAFT~\cite{teed2020raft} and GMFlow~\cite{xu2022gmflow} provide dense pixel-wise motion cues. 
Recent feed-forward 4D methods further exploit flow or scene flow for dynamic supervision~\cite{cong2026flow3r}. 
However, such motion cues are often formulated as standalone output-level objectives, leaving their relation to the predicted scene geometry and temporal state evolution only implicitly modeled. 
In contrast, our framework introduces Unified Motion Supervision (UMS), which couples inter-frame motion with evolving 3D geometry and provides geometry-grounded supervision for temporally coherent state evolution.

\section{Method}
\label{sec:method}

\paragraph{Problem Formulation.}
We formulate monocular dynamic 4D visual geometry learning as feed-forward prediction over an ordered video stream.
Let the input be a sequence of \(T\) frames, 
\(\mathcal{I}=(\mathbf{I}_{1},\ldots,\mathbf{I}_{T})\), where each frame 
\(\mathbf{I}_{t}\in\mathbb{R}^{H\times W\times3}\).
The network \(f_{\theta}\) maps this ordered sequence to a tuple of temporally aligned camera, geometry, and motion outputs:
\begin{equation}
f_{\theta}(\mathcal{I}) =
\left(
(\hat{\mathbf{T}}_1,\ldots,\hat{\mathbf{T}}_T),
(\hat{\mathbf{K}}_1,\ldots,\hat{\mathbf{K}}_T),
(\hat{\mathbf{X}}_1,\ldots,\hat{\mathbf{X}}_T),
(\hat{\mathbf{F}}_1,\ldots,\hat{\mathbf{F}}_{T-1})
\right).
\label{eq:problem_formulation}
\end{equation}
Here, \(\hat{\mathbf{T}}_{t}\in\mathbb{R}^{4\times4}\) denotes the camera extrinsic matrix,
\(\hat{\mathbf{K}}_{t}\in\mathbb{R}^{3\times3}\) denotes the camera intrinsic matrix,
\(\hat{\mathbf{X}}_{t}\in\mathbb{R}^{H\times W\times3}\) denotes the pixel-aligned 3D point map for frame \(\mathbf{I}_{t}\),
and \(\hat{\mathbf{F}}_{t}\in\mathbb{R}^{H\times W\times2}\) denotes the forward optical flow from \(\mathbf{I}_{t}\) to \(\mathbf{I}_{t+1}\).
Since monocular 4D visual geometry is defined over an ordered video sequence, the temporal order of frames specifies the direction along which scene states evolve.

\begin{figure}[t]
    \centering
    \includegraphics[width=\linewidth]{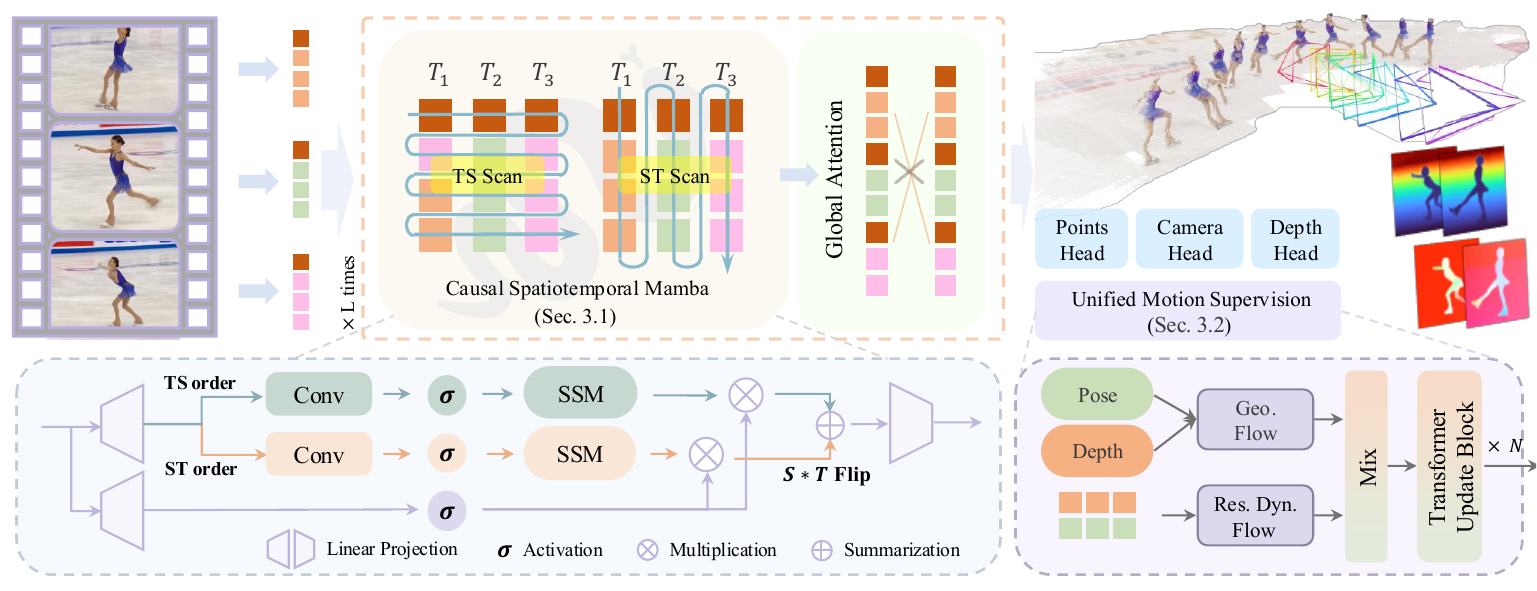}
    \caption{
    \textbf{Overview of our causal spatiotemporal state evolution framework.}
    Given an ordered monocular video, the model first extracts geometry-aware tokens with global context, and then applies {Causal Spatiotemporal Mamba (CSM)} to propagate these tokens through causal spatiotemporal state evolution.
    Complementary \textit{Temporal-Spatial Scan} (TS Scan) and \textit{Spatial-Temporal Scan} (ST Scan) update geometry-aware states across temporal and spatial dimensions while preserving the temporal order of scene evolution.
    The updated spatiotemporal representations parameterize camera extrinsics, intrinsics, pixel-aligned 3D point maps, and forward optical flow.
    {Unified Motion Supervision (UMS)} constructs a unified forward motion field with geometry-grounded supervision mechanism, enforcing geometry-aligned inter-frame state transitions.
    }
\label{fig:CSM}
\end{figure}


\paragraph{Overall Architecture.}
Under this formulation, our framework realizes monocular 4D visual geometry learning through causal spatiotemporal state evolution, as shown in \Cref{fig:CSM}. 
It consists of two main components:
(1) \textit{Causal Spatiotemporal Mamba (CSM)} propagates geometry-aware tokens with directional state updates, allowing temporally ordered representation evolution rather than relying solely on correlation-based temporal aggregation (Section~\ref{sec:CSM}); 
(2) \textit{Unified Motion Supervision (UMS)} constructs a geometry-grounded unified forward motion field from camera, geometry, and residual motion predictions, coupling inter-frame motion with evolving 3D geometry to guide temporally coherent state evolution (Section~\ref{sec:UMS}). 
The resulting spatiotemporal representations parameterize the camera, geometry, and motion-related outputs defined in Equation~\eqref{eq:problem_formulation}.

\subsection{Causal Spatiotemporal Mamba}
\label{sec:CSM}
\begin{figure}[t]
    \centering
    \includegraphics[width=\linewidth]{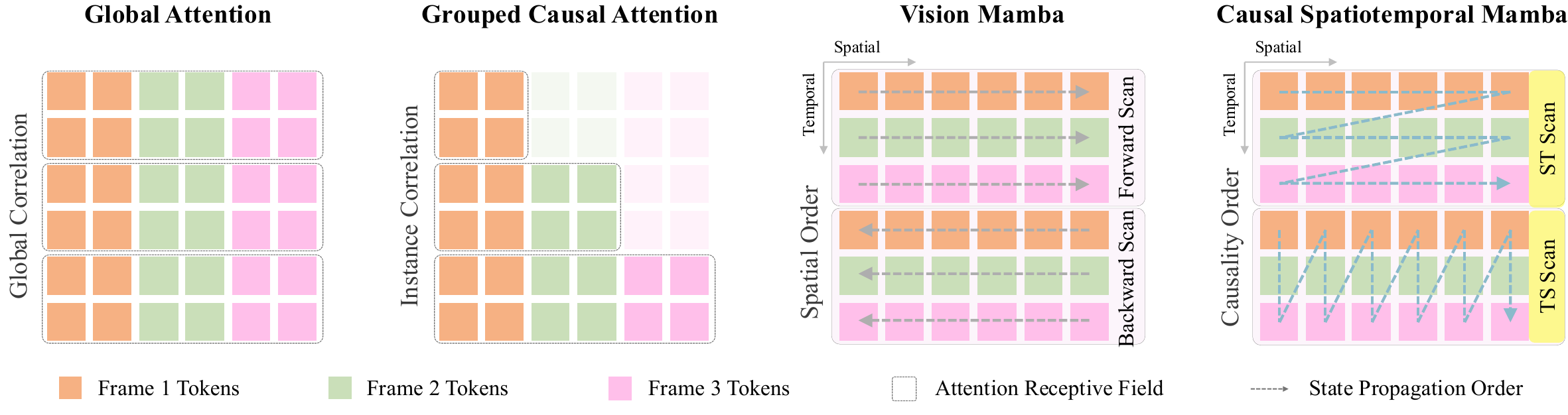}
\caption{
\textbf{Comparison of  4D modeling paradigms.}
Attention-based methods~\cite{wang2025vggt,zhou2026page4d} aggregate context through pairwise token affinities, while Image/Video Mamba~\cite{liu2024vmamba,li2024videomamba} applies generic bidirectional scans for visual representation learning.
In contrast, CSM introduces geometry-aware causal scans that propagate states along the temporal order of the video.
The complementary \textit{Temporal-Spatial Scan} and \textit{Spatial-Temporal Scan} jointly model temporal evolution and spatial structure, enabling ordered spatiotemporal state propagation for monocular 4D visual geometry.
}
    \label{fig:Mamba}
\end{figure}

Monocular 4D visual geometry benefits from both global geometric context and temporally ordered state evolution. 
While global attention is effective for long-range token aggregation, it does not explicitly propagate geometry-aware states along the video order. 
We introduce CSM to augment geometry-aware tokens with directional state-space propagation. 
Unlike generic visual Mamba designs for image or video recognition~\cite{liu2024vmamba,li2024videomamba}, CSM adopts structured spatiotemporal scans tailored to 4D geometry prediction, preserving intra-frame spatial structure while modeling inter-frame temporal evolution, as shown in Figure~\ref{fig:CSM} and Figure~\ref{fig:Mamba}.

\paragraph{From Attention Aggregation to State-Space Propagation.}
For a flattened spatiotemporal token \(\mathbf{g}_i\), attention-based aggregation retrieves information from a visible token set \(\mathcal{V}(i)\):
\begin{equation}
\mathbf{y}_i
=
\sum_{j\in\mathcal{V}(i)}
\alpha_{ij}\mathbf{W}_v\mathbf{g}_j,
\qquad
\alpha_{ij}
=
\operatorname{softmax}_{j}
\left(
\frac{
(\mathbf{W}_q\mathbf{g}_i)^{\top}
(\mathbf{W}_k\mathbf{g}_j)
}{\sqrt{d}}
\right),
\label{eq:attention_pooling}
\end{equation}
where \(\mathcal{V}(i)\) denotes the visible token set.
Global attention sets \(\mathcal{V}(i)\) to all video tokens, while causal or grouped variants restrict it using temporal masks or local groups.
In all cases, the update is computed as a weighted aggregation over pairwise token affinities.
Such a mechanism specifies which tokens contribute to the output, but does not explicitly define a state transition process for geometry-aware representations along the temporal order of the video, as shown in Figure~\ref{fig:Mamba}.

CSM therefore augments global correlation modeling with state-space propagation over serialized geometry-aware tokens.
Given a scan order \(r\), let
\(\mathbf{S}^{r}=\{\mathbf{s}^{r}_{i}\}_{i=1}^{TN}\)
denote the token sequence obtained by serializing the spatiotemporal token grid according to \(r\), where \(T\) is the number of frames and \(N\) is the number of spatial tokens per frame.
Starting from \(\mathbf{h}^{r}_{0}=\mathbf{0}\), the state-space update is
\begin{equation}
\mathbf{h}^{r}_{i}
=
\bar{\mathbf{A}}^{r}_{i}\mathbf{h}^{r}_{i-1}
+
\bar{\mathbf{B}}^{r}_{i}\mathbf{s}^{r}_{i},
\qquad
\mathbf{y}^{r}_{i}
=
\mathbf{C}^{r}_{i}\mathbf{h}^{r}_{i}.
\label{eq:csm_recurrence}
\end{equation}
Unrolling Equation~\eqref{eq:csm_recurrence} yields
\begin{equation}
\mathbf{y}^{r}_{i}
=
\sum_{j=1}^{i}
\mathbf{K}^{r}_{i,j}\mathbf{s}^{r}_{j},
\qquad
\mathbf{K}^{r}_{i,j}
=
\mathbf{C}^{r}_{i}
\left(
\prod_{k=j+1}^{i}
\bar{\mathbf{A}}^{r}_{k}
\right)
\bar{\mathbf{B}}^{r}_{j},
\label{eq:scan_kernel}
\end{equation}
Here, \(\mathbf{K}^{r}_{i,j}\) denotes the effective transition from position \(j\) to position \(i\) along scan order \(r\), with the empty product defined as identity. 
Compared with attention-based aggregation, state-space propagation accumulates context through recursively composed state transitions over the serialized token sequence. 
This makes the directional dependency along the scan explicit and provides an ordered propagation mechanism for geometry-aware spatiotemporal representations.


\paragraph{Temporal-Spatial and Spatial-Temporal Scans.}
As shown in Figure~\ref{fig:CSM}, given geometry-aware tokens 
\(\mathbf{G}\in\mathbb{R}^{B\times T\times N\times C}\), 
where \(T\) is the number of frames and \(N\) is the number of spatial tokens per frame, we denote the \(n\)-th token in frame \(t\) as \(\mathbf{g}_{t,n}\).
CSM serializes the same token grid with two complementary scan orders.
The \textit{Temporal-Spatial Scan} (TS Scan) prioritizes temporal trajectories by scanning tokens with the same spatial index consecutively:
\begin{equation}
\mathbf{S}^{\mathrm{TSS}}
=
[
\mathbf{g}_{1,1},\ldots,\mathbf{g}_{T,1},
\ldots,
\mathbf{g}_{1,N},\ldots,\mathbf{g}_{T,N}
].
\label{eq:tss_order}
\end{equation}
The \textit{Spatial-Temporal Scan} (ST Scan) instead follows a frame-major order, aggregating spatial context within each frame before advancing in time:
\begin{equation}
\mathbf{S}^{\mathrm{STS}}
=
[
\mathbf{g}_{1,1},\ldots,\mathbf{g}_{1,N},
\ldots,
\mathbf{g}_{T,1},\ldots,\mathbf{g}_{T,N}
].
\label{eq:sts_order}
\end{equation}
Both sequences are processed by the state-space update in \Cref{eq:csm_recurrence}, reshaped back to the original layout, and fused by
\begin{equation}
\tilde{\mathbf{G}}
=
\Phi_{\mathrm{fuse}}
\left(
\mathbf{Y}^{\mathrm{TSS}},
\mathbf{Y}^{\mathrm{STS}}
\right),
\label{eq:tss_sts_fusion}
\end{equation}
where \(\mathbf{Y}^{\mathrm{TSS}}\) and \(\mathbf{Y}^{\mathrm{STS}}\) are the two scan outputs,$\Phi_{\mathrm{fuse}}\left(\cdot\right)$ denotes a linear projection layer and $\tilde{\mathbf{G}}$ is the output of CSM block.
Together, this dual-scan design ensures that TS Scan models how regions evolve over time, while ST Scan constructs the geometric layout of each frame.

\subsection{Unified Motion Supervision}
\label{sec:UMS}

While CSM propagates geometry-aware states along the temporal order, it does not by itself guarantee that adjacent-frame transitions remain consistent with the underlying 3D scene structure. To address this, we introduce Unified Motion Supervision (UMS), which enforces geometry-consistent state evolution by grounding inter-frame transitions in motion signals derived from predicted camera and depth. Specifically, UMS constructs a unified forward motion field
$\hat{\mathbf{F}}_{t}
=
\hat{\mathbf{F}}^{\mathrm{geo}}_{t}
+
\hat{\mathbf{F}}^{\mathrm{dyn}}_{t},$
where $\hat{\mathbf{F}}^{\mathrm{geo}}_{t}$ is induced by projective geometry using predicted depth, intrinsics, and relative pose, and $\hat{\mathbf{F}}^{\mathrm{dyn}}_{t}$ captures complementary motion components not explained by the geometry-induced term. By supervising this composed motion field, UMS enforces that state transitions are both temporally ordered and geometrically valid, effectively anchoring spatiotemporal representation dynamics to a physically consistent 3D scene.

\paragraph{Composition of Forward Motion Field.}
For each adjacent frame pair \((\mathbf{I}_t,\mathbf{I}_{t+1})\), UMS first derives a geometry-induced motion field from the predicted geometry and camera motion. 
For a pixel \(\mathbf{p}_t=(u,v,1)^\top\), we project its depth-backprojected 3D point to the next frame:
\begin{equation}
\mathbf{p}_{t+1}
=
\pi\!\left(
\hat{\mathbf{K}}_{t+1}
\left(
\hat{\mathbf{R}}_{t\rightarrow t+1}
\hat{\mathbf{D}}_t(u,v)
\hat{\mathbf{K}}_t^{-1}\mathbf{p}_t
+
\hat{\mathbf{t}}_{t\rightarrow t+1}
\right)
\right),
\qquad
\hat{\mathbf{F}}^{\mathrm{geo}}_{t}(u,v)
=
\mathbf{p}_{t+1}
-
\mathbf{p}_t .
\label{eq:ego_flow}
\end{equation}
Here, \(\pi(\cdot)\) denotes perspective projection, and the camera parameters are obtained from the predicted intrinsics and relative pose. 
This $\hat{\mathbf{F}}^{\mathrm{geo}}_{t}$ explains the displacement induced by camera motion under the predicted scene geometry.

To capture motion not explained by geometry-induced term, UMS predicts an complementary dynamic components from both geometry-aware states and image features. 
Specifically, we concatenate the updated geometry state \(\tilde{\mathbf{G}}_t\) with the corresponding image feature \(\mathbf{R}_t\) to form \(\mathbf{Z}_t=[\tilde{\mathbf{G}}_t,\mathbf{R}_t]\). 
Using \(\hat{\mathbf{F}}^{\mathrm{geo}}_t\) as a geometry-derived initialization, the fused feature of frame \(t+1\) is warped toward frame \(t\), and encoded by a learnable predictor \(\psi_{\phi}\):
\begin{equation}
\hat{\mathbf{F}}^{\mathrm{dyn}}_t
=
\psi_{\phi}
\left(
\mathbf{Z}_t,
\mathcal{W}(\mathbf{Z}_{t+1},\hat{\mathbf{F}}^{\mathrm{geo}}_t),
\hat{\mathbf{F}}^{\mathrm{geo}}_t
\right),
\label{eq:object_flow}
\end{equation}
where \(\mathcal{W}(\cdot)\) denotes differentiable backward warping. The predictor \(\psi_{\phi}\) follows the design style of WAFT~\cite{wang2026waft}, but remove the DPT head and operates on our spatiotemporal representations.
The final forward flow and UMS objective are:
\begin{equation}
\hat{\mathbf{F}}_t
=
\hat{\mathbf{F}}^{\mathrm{geo}}_t
+
\hat{\mathbf{F}}^{\mathrm{dyn}}_t,
\qquad
\mathcal{L}_{\mathrm{UMS}}
=
\sum_{t=1}^{T-1}
\rho\!\left(
\hat{\mathbf{F}}_t-\mathbf{F}^{\ast}_t
\right),
\label{eq:ums_loss}
\end{equation}
where \(\mathbf{F}^{\ast}_t\) is the optical-flow target and \(\rho(\cdot)\) is a robust penalty.
By supervising the composed motion field, UMS enforces consistency between geometry predictions and inter-frame state transitions, providing unified geometry-aware motion supervision for causal state evolution.

\subsection{Optimization}

The proposed framework is trained end-to-end with joint geometry and motion supervision. For geometry supervision, we adopt the standard geometric losses used in prior feed-forward reconstruction methods~\cite{wang2025pi}, supervising camera poses, depth maps, and pixel-aligned 3D point maps with scale alignment. In addition to geometric supervision, we further introduce UMS as an additional motion constraint. For motion supervision, we supervise the forward flow composed by UMS using dense optical-flow targets.
The overall objective is
\begin{equation}
\mathcal{L}
=
\lambda_{\mathrm{cam}}\mathcal{L}_{\mathrm{cam}}
+
\lambda_{\mathrm{depth}}\mathcal{L}_{\mathrm{depth}}
+
\lambda_{\mathrm{point}}\mathcal{L}_{\mathrm{point}}
+
\lambda_{\mathrm{UMS}}\mathcal{L}_{\mathrm{UMS}},
\end{equation}
where \(\mathcal{L}_{\mathrm{cam}}\) supervises camera translation and rotation, 
\(\mathcal{L}_{\mathrm{depth}}\) and \(\mathcal{L}_{\mathrm{point}}\) supervise scale-aligned depth and 3D point predictions, 
and \(\mathcal{L}_{\mathrm{UMS}}\) supervises the composed forward flow and visibility predictions.

\section{Experiments}
\label{sec:eval}

\paragraph{Implementation Details.}
Our framework builds upon a VGGT-style architecture~\cite{wang2025vggt}. 
We use a frozen DINOv2 ViT-L/14 backbone with register tokens to extract dense geometric features, and construct the 4D representation with a 36-layer decoder that alternates between causal Mamba sequence modeling and global spatiotemporal attention. 
During training, we randomly sample contiguous clips of 2 to 24 frames with dynamic temporal batching. 
Training is performed in two stages, from low-resolution temporal initialization to high-resolution fine-tuning. 
We use AdamW with a weight decay of 0.05 and a OneCycleLR schedule with a peak learning rate of \(5 \times 10^{-5}\), including a 10\% warm-up followed by cosine annealing.
For evaluation, we follow the protocols and metrics of VGGT~\cite{wang2025vggt} and CUT3R~\cite{wang2025cut3r}. 
We compare with VGGT~\cite{wang2025vggt}, StreamVGGT~\cite{zhuo2026streamvggt}, MoRe~\cite{fang2026more}, PAGE-4D~\cite{zhou2026page4d}, and CUT3R~\cite{wang2025cut3r} using their publicly released pretrained models. 
Additional architectural and training details are provided in the Appendix.


\paragraph{Training dataset.}
For spatiotemporal adaptation, we construct a unified training corpus from nine datasets: TarTanAir~\cite{wang2020tartanair}, Virtual KITTI2~\cite{cabon2020virtual}, Spring~\cite{mehl2023spring}, PointOdyssey~\cite{zheng2023pointodyssey}, BEDLAM~\cite{black2023bedlam}, DynamicReplica~\cite{karaev2023dynamicstereo}, GTA-SfM~\cite{huang2018deepmvs}, Hypersim~\cite{Hypersim}, and ARKitScenes~\cite{baruch2021arkitscenes}. These datasets cover diverse scene types and motion patterns, including autonomous driving, indoor and outdoor environments, articulated human motion, synthetic object dynamics, and handheld camera trajectories.

\paragraph{Evaluation datasets.}
We evaluate our model across three geometry and motion perception tasks: camera pose estimation, video depth estimation, and optical flow estimation.
Our evaluation suite includes Sintel~\cite{butler2012naturalistic}, Bonn~\cite{palazzolo2019refusion}, and KITTI~\cite{menze2015object}, covering synthetic and real-world scenes, indoor and outdoor environments, and a range of dynamic motion patterns.
For video depth estimation, we report results on Sintel~\cite{butler2012naturalistic}, Bonn~\cite{palazzolo2019refusion}, and KITTI~\cite{menze2015object}.
For camera pose estimation, we use Sintel~\cite{butler2012naturalistic}, and Bonn~\cite{palazzolo2019refusion}, as they contain challenging dynamic motions and viewpoint changes that stress-test temporal geometric consistency.
Finally, we evaluate optical flow estimation on the Sintel~\cite{butler2012naturalistic} and KITTI~\cite{menze2015object} datasets following the standard evaluation protocol, where dense public ground truth is available.

\begin{table*}[t]
\centering
\small
\caption{Camera pose estimation on dynamic benchmarks. Best results are in \textbf{bold}, second-best are \underline{underlined}.}
\label{tab:pose_estimation}
\begin{tabular}{lcccccc}
\toprule
\multirow{2}{*}{\textbf{Method}} 
& \multicolumn{3}{c}{\textbf{Sintel}} 
& \multicolumn{3}{c}{\textbf{Bonn}} \\
\cmidrule(lr){2-4}\cmidrule(lr){5-7}
& ATE$\downarrow$ & RPE-T$\downarrow$ & RPE-R$\downarrow$ 
& ATE$\downarrow$ & RPE-T$\downarrow$ & RPE-R$\downarrow$ \\
\midrule
VGGT       & 0.214 & 0.079 & 0.643 & \underline{0.021} & 0.023 & 0.937 \\
StreamVGGT & 0.216 & 0.106 & 0.779 & 0.045 & 0.010 & \underline{0.662} \\
MoRe       & \underline{0.147} & 0.082 & 0.616 & \underline{0.021} & 0.012 & \textbf{0.650} \\
PAGE-4D    & 0.175 & 0.074 & \underline{0.602} & \underline{0.021} & 0.013 & \textbf{0.650} \\
CUT3R      & 0.235 & \underline{0.068} & 0.669 & 0.035 & \underline{0.009} & 0.680 \\
\rowcolor{gray!12}
\textbf{Ours}  & \textbf{0.116} & \textbf{0.050} & \textbf{0.552} & \textbf{0.020} & \textbf{0.008} & 0.670 \\
\bottomrule
\end{tabular}
\end{table*}

\subsection{Results}

\paragraph{Monocular Video Depth Estimation} We evaluate monocular video depth estimation using Absolute Relative Error (Abs Rel) and threshold accuracy \(\delta < 1.25\), where lower Abs Rel and higher accuracy indicate better depth prediction. 
For fair comparison across methods with different depth parameterizations, we apply a unified alignment protocol before evaluation. 
Specifically, disparity or affine-invariant relative depth predictions are aligned to ground-truth depth using median scaling and scale-and-shift alignment.
Table~\ref{tab:depth_estimation} reports results on Sintel, Bonn, and KITTI. 
Our model performs consistently across synthetic, real-world, indoor, and outdoor scenes. 
The improvements suggest that the proposed spatiotemporal representation effectively exploits cross-frame geometric cues, improving temporal depth consistency while maintaining accurate per-frame geometry.

\paragraph{Camera Pose Estimation}
We evaluate camera pose estimation on Sintel, and Bonn using absolute trajectory error (ATE), translational relative pose error (RPE-T), and rotational relative pose error (RPE-R), where lower values indicate better pose recovery.
Table~\ref{tab:pose_estimation} reports the results.
Compared with prior feed-forward 4D methods, our model shows improved pose accuracy in challenging dynamic scenes, including large viewpoint changes on Sintel.
These gains suggest that causal spatiotemporal state propagation helps maintain temporally coherent pose estimates, while UMS couples inter-frame motion with evolving 3D geometry to improve trajectory consistency.

\begin{table*}[t]
\centering
\small
\caption{Video depth estimation on dynamic benchmarks. Metric $\delta < 1.25$ is reported in \%. Best results are in \textbf{bold}, and second-best results are \underline{underlined}. ``OOM'' denotes out-of-memory under the same single-pass feed-forward inference setting.}
\label{tab:depth_estimation}
\begin{tabular}{lcccccc}
\toprule
\multirow{2}{*}{\textbf{Method}} 
& \multicolumn{2}{c}{\textbf{Sintel}} 
& \multicolumn{2}{c}{\textbf{Bonn}} 
& \multicolumn{2}{c}{\textbf{KITTI}} \\
\cmidrule(lr){2-3}\cmidrule(lr){4-5}\cmidrule(lr){6-7}
& Abs Rel$\downarrow$ & $\delta < 1.25 \uparrow$ 
& Abs Rel$\downarrow$ & $\delta < 1.25 \uparrow$ 
& Abs Rel$\downarrow$ & $\delta < 1.25 \uparrow$ \\
\midrule
VGGT & 0.471 & 0.486 & 0.103 & 0.903 & 0.143 & 0.804\\
StreamVGGT & 0.354 & 0.591 & OOM & OOM & 0.176 & 0.718 \\
MoRE & 0.340 & \underline{0.592}  & \textbf{0.057} & \underline{0.967} & 0.141 & 0.878\\
PAGE-4D & \underline{0.299} & \textbf{0.735} & 0.087 & 0.907 & 0.147 & \underline{0.894} \\

CUT3R & 0.534 & 0.558 & 0.075 & 0.943 & \underline{0.118} & 0.881 \\
\rowcolor{gray!12}
\textbf{Ours} & \textbf{0.243} & \textbf{0.735} & \underline{0.059} & \textbf{0.972} & \textbf{0.086} & \textbf{0.937} \\

\bottomrule
\end{tabular}
\end{table*}

\begin{figure}[t]
    \centering
    \includegraphics[width=\linewidth]{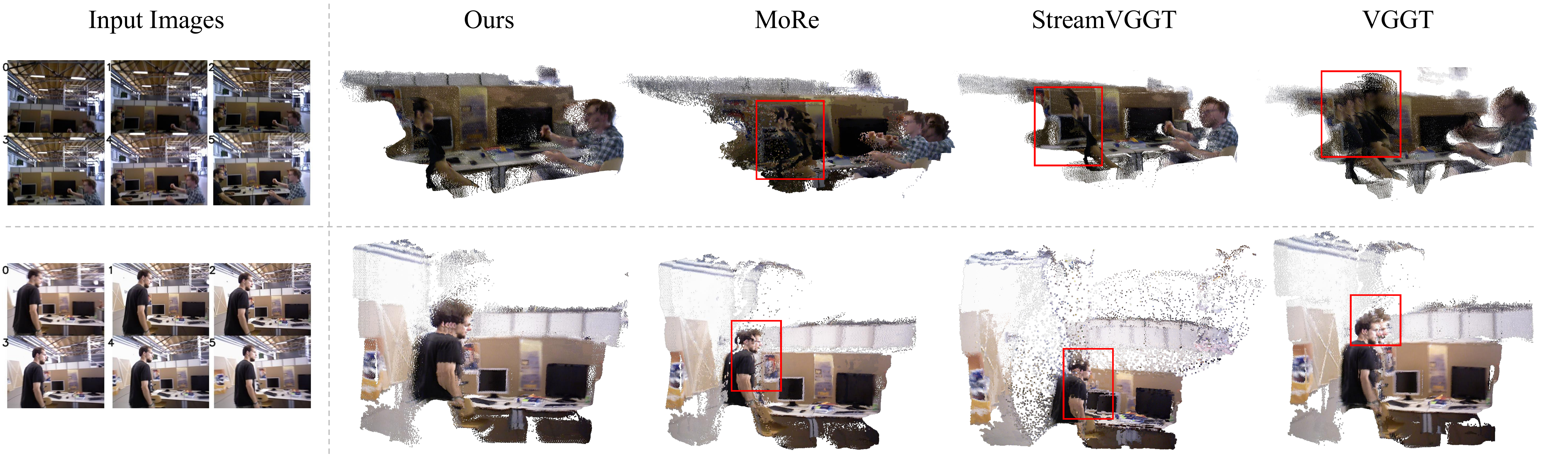}
    \caption{Qualitative comparison of dynamic geometry reconstruction. Our method produces more accurate and temporally coherent geometry in dynamic regions, with sharper structures and fewer artifacts than existing feed-forward 4D methods.}
    \label{fig:show}
\end{figure}

\paragraph{Optical Flow Estimation}

\begin{figure}[htbp]
    \centering
    \includegraphics[width=\linewidth]{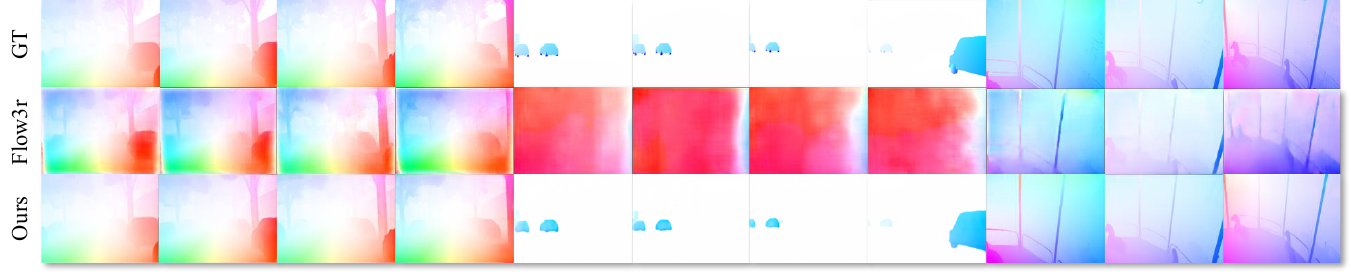}
    \caption{Qualitative comparison of optical flow estimation. Benefiting from our unified motion supervision, our model yields highly accurate motion fields with sharp boundaries. Notably, in the middle columns where moving objects occupy a small proportion and share high visual similarity with the background, Flow3r struggles to effectively detect the foreground optical flow. In contrast, our approach successfully disentangles these challenging motions.}
    \label{fig:flow}
\end{figure}
Although our framework is primarily designed for geometry-centric 4D scene understanding, evaluating the derived optical flow provides an additional way to assess the geometry-motion coupling induced by UMS. 
Specifically, we evaluate optical flow estimation on the Sintel and KITTI splits using the standard End-Point Error (EPE) metric and compare against the state-of-the-art feed-forward optical flow method Flow3r~\cite{cong2026flow3r}. 
This evaluation examines whether the proposed directional state propagation and unified motion supervision can produce dense motion correspondences that are consistent with dynamic scene geometry.
As illustrated qualitatively in Figure~\ref{fig:flow} and reported quantitatively in Table~\ref{tab:optical_flow}, our model achieves strong optical flow performance.  
\begin{wraptable}{r}{0.3\textwidth}
\vspace{-0.5em}
\centering
\scriptsize
\setlength{\tabcolsep}{1.7pt}
\caption{Comparison results of optical flow estimation.}
\label{tab:optical_flow}
\begin{tabular}{lcc}
\toprule
\textbf{Method} & \textbf{Sintel EPE$\downarrow$} & \textbf{KITTI EPE$\downarrow$} \\
\midrule
Flow3r & 45.146 & 38.649 \\
\rowcolor{gray!12}
\textbf{Ours}  & \textbf{16.108} & \textbf{20.161} \\
\bottomrule
\end{tabular}
\vspace{-1em}
\end{wraptable}
These results indicate that UMS helps align spatiotemporal state propagation with observable inter-frame motion. 
Rather than treating optical flow as an isolated output task, our framework derives motion fields within a geometry-aware formulation, providing complementary evidence that the learned spatiotemporal representations capture meaningful dynamic scene evolution. 
This geometry-motion coupling also supports the improvements observed in video depth and camera pose estimation.

\paragraph{Qualitative Results.}
As shown in~\Cref{fig:flow}, the visualization results demonstrate that our method produces noticeably better dynamic geometry than existing feed-forward 4D methods. 
In challenging regions~\cite{sturm2012benchmark} involving object motion, large deformation, or camera-induced motion, competing methods often suffer from distorted object shapes, blurred structures, unstable depth boundaries, or temporally inconsistent geometry across adjacent frames. 
In contrast, our model better preserves object contours and maintains more coherent 3D structures over time, leading to sharper and more stable reconstructions of dynamic scenes. 
This improvement suggests that the proposed causal spatiotemporal state propagation helps maintain temporally ordered geometry evolution, while Unified Motion Supervision provides geometry-grounded motion guidance for dynamic regions. 
Together, these qualitative results further highlight the benefit of coupling directional state propagation with unified motion supervision for geometry-aware dynamic scene modeling.

\subsection{Ablation Study}
\label{ablation}

We ablate the key design choices on Sintel, as shown in \Cref{tab:ablation}. 
For a controlled comparison, all variants are trained for 100 epochs on the same training subset. 
We compare five configurations: VGGT-style global attention only, causal attention, Vision-Mamba-style modeling extended to video~\cite{zhu2024vision}, our CSM, and the full model with UMS.

\textbf{Effect of spatiotemporal modeling.}
As shown in Table~\ref{tab:abl} rows 1 -- 4 compare different spatiotemporal modeling paradigms. 
VGGT-style Global attention only captures long-range correlations but does not explicitly encode temporal order. 
Causal attention improves temporal directionality through masking, but still relies on pairwise token aggregation rather than state propagation. 
Vision-Mamba-style modeling introduces state-space propagation, yet its generic scan pattern is not tailored to geometry-aware video modeling. 
In contrast, CSM achieves better performance across depth and pose metrics, suggesting that the proposed Temporal-Spatial and Spatial-Temporal scans more effectively capture ordered temporal evolution while preserving spatial geometric structure.

\textbf{Effect of Unified Motion Supervision.}
Adding UMS on top of CSM further improves both video depth and pose estimation. 
This indicates that directional state propagation alone benefits from additional geometry-grounded motion supervision. 
By coupling inter-frame motion with predicted geometry, UMS provides a structural constraint for learning more temporally coherent representations, leading to reduced trajectory error and improved depth consistency.
Overall, the ablation confirms that CSM and UMS provide complementary benefits: CSM improves causal spatiotemporal representation learning, while UMS further strengthens geometry-motion consistency.

\begin{wraptable}{r}{0.75\textwidth}
\vspace{-0.5em}
\centering
\scriptsize
\setlength{\tabcolsep}{2.5pt}
\caption{Ablation study on architectural components and training strategies. Evaluation is performed on Sintel. Best results are shown in bold.}
\label{tab:ablation}
\resizebox{\linewidth}{!}{%
\begin{tabular}{cc|cc|ccc}
\toprule
\multicolumn{2}{c|}{Components}
& \multicolumn{2}{c|}{Video Depth}
& \multicolumn{3}{c}{Pose} \\
\cmidrule(lr){1-2}
\cmidrule(lr){3-4}
\cmidrule(lr){5-7}
Modeling & UMS
& Abs Rel$\downarrow$ & $\delta < 1.25 \uparrow$
& ATE$\downarrow$ & RPE-T$\downarrow$ & RPE-R$\downarrow$ \\
\midrule
Only Global-Attn & \xmark
& 0.481 & 0.500
& 0.321 & 0.213 & 0.814 \\
Causal-Attn & \xmark
& 0.451 & 0.592 
& 0.311 & 0.191 & 0.637 \\
Vision Mamba & \xmark
& 0.309 & 0.564
& 0.329 & 0.177 & 0.732 \\
\textbf{CSM} & \xmark
& 0.253 & 0.631
& 0.124 & 0.078 & 0.559 \\
\rowcolor{gray!12}
\textbf{CSM} & \cmark
& \textbf{0.243} & \textbf{0.689}
& \textbf{0.116} & \textbf{0.050} & \textbf{0.552} \\
\bottomrule
\end{tabular}%
}
\label{tab:abl}
\vspace{-1em}
\end{wraptable}


\section{Conclusion}
\label{sec:conclusion}
In this paper, we formulated monocular 4D visual geometry learning as causally ordered spatiotemporal state evolution, providing a structured perspective for extending feed-forward 3D reconstruction to dynamic scenes. 
We introduced Causal Spatiotemporal Mamba (CSM), which augments geometry-aware representations with directional state propagation, enabling temporally ordered spatiotemporal modeling beyond purely correlation-based temporal aggregation. 
Furthermore, our Unified Motion Supervision (UMS) provides a geometry-grounded approach to motion supervision. 
By constructing a unified motion field from predicted camera, geometry, and residual motion cues, UMS couples inter-frame motion with evolving 3D structure and provides a structural constraint for temporally coherent state evolution. 
Extensive evaluations demonstrate the effectiveness of our approach, showing strong performance in dynamic geometry estimation and improved temporal consistency across diverse scenes.

\paragraph{Limitations and Future Work.}
While our framework improves dynamic scene representation, UMS relies on reasonably accurate camera and depth predictions to derive geometry-consistent motion signals. 
In scenarios involving severe motion blur, degenerate camera trajectories, large occlusions, or unreliable geometry estimates, the resulting motion supervision may become less effective. 
Future work could integrate learnable pose-refinement modules into the causal spatiotemporal latent space, enabling joint refinement of camera trajectories and geometry-aware states. 
We also plan to investigate the scalability of the proposed state-propagation paradigm on larger and more heterogeneous in-the-wild video datasets. 
More broadly, this work suggests that aligning sequence modeling with geometry-aware motion constraints is a promising direction for scalable and temporally consistent 4D visual representation learning.

{
\small
\bibliographystyle{plain}
\bibliography{main}
}






\end{document}